# A stochastic model for Case-Based Reasoning

Michael Gr. Voskoglou
Graduate Technological Educational Institute of Patras, Greece
voskoglou@teipat.gr , mvosk@hol.gr

**Abstract**
Case-Based Reasoning (CBR) is the process of solving new problems based on the solution of similar past problems. In the present paper we introduce an absorbing Markov chain on the main steps of the CBR process. In this way we succeed in obtaining the probabilities for the above process to be in a certain step at a certain phase of the solution of the corresponding problem, and a measure for the efficiency of a CBR system. Examples are also given to illustrate our results.

**Keywords:** Case-Based Reasoning, Problem-Solving, Finite Markov Chains, Intelligent Systems

## 1. Introduction

Case-Based Reasoning (CBR) is a recent theory for problem-solving and learning in computers and people. Broadly construed it is the process of solving new problems based on the solutions of similar past problems. The term problem-solving is used here in a wide sense, which means that it is not necessarily the finding of a concrete solution to an application problem, it may be any problem put forth by the user. For example to justify, or criticize a proposed solution, to interpret a problem situation, to generate a set of possible solutions, or generate explanations in observable data, are also problem solving situations. A lawyer, who advocates a particular outcome in a trial based on legal precedents, or an auto mechanic, who fixes an engine by recalling another car that exhibited similar symptoms, are using CBR; in other words CBR is a prominent kind of analogy making.

Its coupling to learning occurs as a natural by-product of problem solving. When a problem is successfully solved, the experience is retained in order to solve similar problems in future. When an attempt to solve a problem fails, the reason for the failure is identified and remembered in order to avoid the same mistake in future. Thus CBR is a cyclic and integrated process of solving a problem, learning from this experience, solving a new problem, etc.

The CBR systems' expertise is embodied in a collection (*library*) of past cases rather, than being encoded in classical rules. Each case typically contains a description of the problem plus a solution and/or the outcomes. The knowledge and reasoning process used by an expert to solve the problem is not recorded, but is implicit in the solution.

CBR traces its roots in Artificial Intelligence to the work of Roger Schank and his students at Yale University – U.S.A. in the early 1980's. Scfhank's model of dynamic memory (Schank, 1982) was the basis of the earliest CBR systems that might be called case-based reasoners, Kolodner's CYRUS (1983) and Lebowitz's IPP (1983).

As an intelligent-systems method CBR has got a lot of attention over the last few years, because it enables the information managers to increase efficiency and reduce cost by substantially automating processes. CBR first appeared in commercial systems in early 1990's and since then has been sued to create numerous applications in a wide range of domains including *diagnosis, help-desk, assessment, decision support, design*, etc. Organizations as diverse as IBM, VISA International, Volkswagen, British Airways and NASA have already made use of CBR in fields like customer support, quality assurance, aircraft maintenance, process planning, and many more that are easily imaginable.

For general facts of the CBR process and methods we refer freely to Voskoglou (2008).

## 2. The main steps of the CBR process

CBR has been formalized for purposes of computer and human reasoning as a four steps process. These steps involve:
- $R_1$: *Retrieve* the most similar to the new problem past case.

- $R_2$: *Reuse* the information and knowledge of the retrieved case for the solution of the new problem.
- $R_3$: *Revise* the proposed solution.
- $R_4$: *Retain* the part of this experience likely to be useful for future problem-solving.

More specifically, the retrieve task starts with the description of the new problem, and ends when a best matching previous case has been found. The reuse of the solution of the retrieved case in the context of the new problem focuses on two aspects: The differences between the past and the current case, and what part of the retrieved case can be transferred to the new case. Usually in non trivial situations part of the solution of the retrieved case cannot be directly transferred to the new case, but requires an adaptation process that takes into account the above differences. Through the revision the solution generated by reuse is tested for success – e.g. by being applied to the real world environment, or to a simulation of it, or evaluated by a specialist – and repaired, if failed. When a failure is encountered, the system can then get a reminding of a previous similar failure and use the failure case in order to improve its understanding of the present failure, and correct it. In other words, there is a transfer from $R_3$ to $R_1$ in this case, and the same circle is repeated again. The revised task can then be retained directly (if the $R_3$ task assures its correctness), or it can be evaluated and repaired again. In the latter case the CBR process remains in fact in step $R_3$ for two successive phases. The final step $R_4$ involves selecting which information from the new case to retain, in what form to retain it, how to index the case for better retrieval in future for similar problems, and how to integrate the new case in the memory structure.

According to the above description the *"flow diagram"* of the CBR process can be represented as shown in figure 1 below.

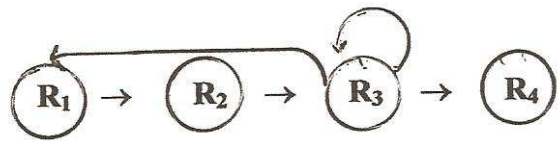

**Figure 1**

Notice that Slade (1991; Figure 1)), Aamodt and Plaza (1994; Figures 1 and 2), Lei et al (2001; Figure 1) and others have presented detailed flowcharts illustrating the basic steps of the CBR process, but in the present paper our target is to achieve a mathematical representation of the CBR process. Towards this direction the brief flow-diagram of Figure 1 will help us to build a stochastic model for the description of the CBR process.

## 3. The stochastic (Markov chain) model

Roughly speaking a *Markov chain* is a stochastic process that moves in a sequence of phases through a set of states and has "no memory". This means that the probability of entering a certain state in a certain phase, although it is not necessarily independent of previous states, depends at most on the state occupied in the previous phase. This property is known as the *Markov property*.

When its set of states is a finite set, then we speak about a *finite Markov chain*. For special facts on such type of chains we refer freely to Kemeny & Snell (1976).

In this paper we present a Markov chain model for the mathematical description of the CBR process. For this, assuming that the CBR process has the Markov property, we introduce a finite Markov chain having as states the four steps of the CBR process described in the previous section. The above assumption is a simplification (not far away from the truth) made to the real system in order to transfer from it to the "assumed real system". This is a standard technique applied during the mathematical modelling process of a real world problem, which enables the formulation of the problem in a form ready for mathematical treatment (see Voskoglou, 2007, section 1).

Denote by $p_{ij}$ the transition probability from state $R_i$ to $R_j$, for i,j=1,2,3,4, then the matrix A=[$p_{ij}$] is said to be the *transition matrix* of the chain.

According to the flow-diagram of the CBR process of Figure 1 we find that

$$A = \begin{array}{c} \\ R_1 \\ R_2 \\ R_3 \\ R_4 \end{array} \begin{array}{cccc} R1 & R2 & R3 & R4 \\ \left[ \begin{array}{cccc} 0 & 1 & 0 & 0 \\ 0 & 0 & 1 & 0 \\ p_{31} & 0 & p_{33} & p_{34} \\ 0 & 0 & 0 & 1 \end{array} \right] \end{array},$$

where we obviously have that $p_{31}+p_{33}+p_{34}=1$ (probability of the certain fact).

Further let us denote by $\varphi_0, \varphi_1, \varphi_2, \ldots\ldots$ the successive phases of the above chain, and also denote by

$$P_i = [p_1^{(i)} p_2^{(i)} p_3^{(i)} p_4^{(i)}]$$

the row - matrix giving the probabilities $p_j^{(i)}$ for the chain to be in each of the states $R_j$, j=1,2,3,4, at phase $\varphi i$, i=1,2,…., where we obviously have again that

$$\sum_{j=1}^{4} p_j^{(i)} = 1.$$

The above row-matrix is called the *probability vector* of the chain at phase $\varphi_i$.

From the transition matrix A and the flow diagram of Figure 1 we obtain the *"tree of correspondence"* among the several phases of the chain and its states shown in Figure 2 below.

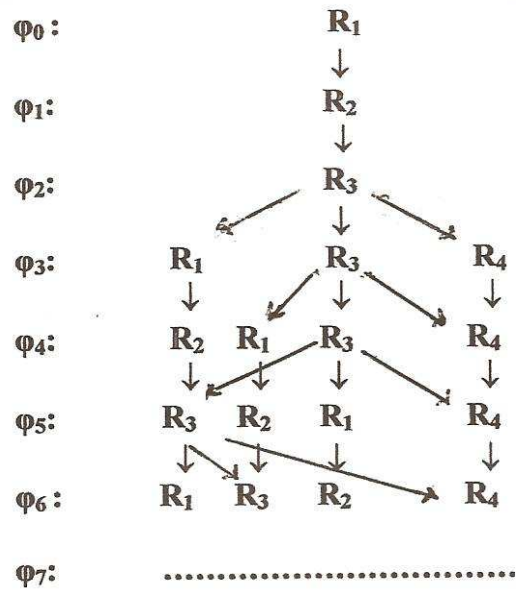

**Figure 2**

From the above tree becomes evident that

$P_0 = [1\ 0\ 0\ 0]$, $P_1 = [0\ 1\ 0\ 0]$, $P_2 = [0\ 0\ 1\ 0]$, and $P_3 = [p_{31}\ 0\ p_{33}\ p_{34}]$.

Further it is well known that

$$P_{i+1} = P_i A, \quad i=0,1,2,\ldots\ldots\ .$$

Therefore we find that

$$P_4 = P_3 A = [p_{33}p_{31} \quad p_{31} \quad p_{33}^2 \quad p_{34}(p_{33}+1)]$$

$$P_5 = P_4 A = [p_{33}^2 p_{31} \quad p_{33}p_{31} \quad p_{31}+p_{33}^3 \quad p_{34}(p_{33}^2+p_{33}+1)]$$

and so on.

Observe now that, when the chain reaches state $R_4$, it is impossible to leave it, because the solution process of the new problem via the CBR approach finishes there. Thus we have an *absorbing Markov chain* with $R_4$ its unique absorbing state. Applying standard techniques from the theory of absorbing chains we bring the transition matrix A to its *canonical (or standard) form* A* by listing the absorbing state first and then partition it as follows:

$$A^* = \begin{array}{c} \\ R_4 \\ - \\ R_1 \\ R_2 \\ R_3 \end{array} \begin{array}{c} R4 \quad\quad R1 \; R2 \; R3 \\ \begin{bmatrix} 1 & | & 0 & 0 & 0 \\ - & - & - & - & - \\ 0 & | & 0 & 1 & 0 \\ 0 & | & 0 & 0 & 1 \\ p_{34} & | & p_{31} & 0 & p_{33} \end{bmatrix} \end{array}.$$

Symbolically we can write

$$A^* = \begin{bmatrix} I & | & 0 \\ - & | & - \\ R & | & Q \end{bmatrix},$$

where Q stands for the transition matrix of the non absorbing states. Then the *fundamental matrix* of the chain is given by

$$N = (I_3-Q)^{-1} = \frac{adj(I_3-Q)}{D(I_3-Q)},$$

where $I_3$ denotes the 3X3 unitary matrix, adj($I_3$-Q) denotes the adjoin matrix and D($I_3$-Q) denotes the determinant of $I_3$-Q. A straightforward calculation gives that

$$N = \frac{1}{1-p_{31}-p_{33}} \begin{bmatrix} 1-p_{33} & 1-p_{33} & 1 \\ -p_{31} & 1-p_{33} & 1 \\ p_{31} & p_{31} & 1 \end{bmatrix} = [n_{ij}]$$

It is well known that the entry $n_{ij}$ of N gives the mean number of times in state $R_j$ when the chain is started in state $R_i$. Therefore, since the present chain is always starting from $R_1$, the sum

$$t = n_{11}+n_{12}+n_{13} = \frac{3-2p_{33}}{1-p_{31}-p_{33}}$$

gives the mean number of phases of the chain before absorption, in other words the mean number of steps for the completion of the CBR process is t+1. It becomes therefore evident that, the bigger is the value of t, the greater is the difficulty encountered for the solution of the given problem via the CBR process. The ideal case is when the CBR process is completed straightforwardly, i.e. without "backwards" from $R_3$ to $R_1$, or "stays" to $R_3$ (see Figure 1). In this case we have that $p_{31}=p_{33}=0$ and $p_{34}=1$, therefore t=3. Thus in general we have that $t \geq 3$.

The following simple example illustrates our results:

EXAMPLE: Consider the case of a physician, who takes into account the diagnosis and treatment of a previous patient having similar symptoms in order to determine the disease and treatment for the patient in front of him. Obviously the physician is using CBR. If the initial treatment fails to improve the health of the patient, then the physician either revises the treatment (stay to $R_3$ for two successive phases), or, in more difficult cases, gets a reminding of a previous similar failure and uses the failure case to improve its understanding of the present failure and correct it (transfer from $R_3$ to $R_1$). The process is completed, when the physician succeeds to cure the patient.

Assume that the recorded statistical data show that the probabilities of a straightforward cure of the patient and of each of the above two reactions of the physician in case of failure are equal to each other. This means that $p_{13}=p_{33}=p_{34}=\frac{1}{3}$ and therefore t=7, i.e. the mean number of steps for the cure of the patient is 8.

Further, one finds that

$$P_3 = [\frac{1}{3} \quad \frac{1}{3} \quad \frac{1}{3}], P_4 = [\frac{1}{9} \quad \frac{1}{3} \quad \frac{1}{9} \quad \frac{4}{9}], P_5 = [\frac{1}{27} \quad \frac{1}{9} \quad \frac{4}{9} \quad \frac{13}{27}]$$

and so on. Observing for example the probability vector $P_5$ one finds that the probability for the CBR process to be at the step of revision ($R_3$) in the 6th phase after its starting is $\frac{4}{9}$, or approximately 44,44%, the corresponding probability to be at the step of retaining the acquired experience ($R_4$) is $\frac{13}{27}$, or approximately 48,15% (in this case it is possible that the CBR process has arrived to the absorbing state $R_4$ in an earlier phase), etc.

Note: Knowing the exact "movements" during the CBR process one can calculate the number of steps needed for the absorption of the chain directly from the flow-diagram of Figure 1. For example, considering the above case of the physician, assume that the initial treatment given to the patient failed to cure him and the physician got a reminding of a similar failure in the past in order to correct it. Assume further that the new treatment didn't give the expected results and the physician revised it again succeeding in this way to cure the patient. Under the above assumptions it is easy from Figure 1 to check that the number of steps needed for the absorption is exactly 8.

**4. Measuring the efficiency of a CBR system**

The challenge in CBR is to come up with methods that are suited for problem-solving and learning in particular subject domains and for particular application environments. In line with the process model described in section 2, core problems addressed by CBR research can be grouped into five areas: Representation of cases, and methods for retrieval, reuse, revision and retaining the acquired experience. A CBR system should support the problems appearing in the above five areas. A good system should support a variety of retrieval mechanisms and allow them to be mixed when necessary. In addition, the system should be able to handle large case libraries with the retrieval time increasing linearly (at worst) with the number of cases.

Let us consider now a CBR system including a library of n recorded past cases and let $t_i$, as it has been calculated in the previous section, be the mean number of steps for the completion of the CBR process for case $c_i$, i=1,2,…,n. Each $t_i$ could be stored in the system's library together with the corresponding case $c_i$. We define then the system's *efficiency*, say t, to be the mean value of the $t_i$'s of its stored cases, i.e. we have that

$$t = \frac{\sum_{i=1}^{n} t_i}{n}.$$

The more problems are solved in future applications through the given system, the bigger becomes the number n of the stored cases in the system's library and therefore the value of t is changing. As n increases it is normally expected that t will decrease, because the values of the $t_i$'s of the new stored cases would be decreasing. In fact, the bigger is n, the better would be the chance of a new case to "fit" well (i.e. to have minor differences) with a known past case, and therefore the less would be the difficulty of solving the corresponding problem via the CBR process. Thus we could say that a CBR system "behaves well" if, when n tends to infinity, then its efficiency t tends to 3.

EXAMPLE: Consider a CBR system that has been designed in terms of *Schank's model of dynamic memory* for the representation of cases (Schank, 1982). The basic idea of this model is to organize specific cases, which share similar properties, under a more general structure called a *generalized episode* (GE). During storing of a new case, when a feature of it matches a feature of an existing past case, a new GE is created. Hence the memory structure of the system is in fact dynamic, in the sense that similar parts of two case descriptions are dynamically generalized in to a new GE and the cases are indexed under this GE by their different features.

In order to calculate the efficiency of a system of this type we need first to calculate the efficiencies of the GE's contained in it. For example, assume that the given system contains a GE including three cases, say $c_1$, $c_2$ and $c_3$. Assume further that $c_1$ corresponds to a straightforward successful application of the CBR process, that $c_2$ is the case described in the example of section 3, and that $c_3$ includes one "return" from $R_3$ to $R_1$ and two "stays" to $R_3$. Then $t_1=3$ and $t_2=7$, while for the calculation of $t_3$ observe that $p_{31}=p_{34}=\frac{1}{4}$ and $p_{33}=\frac{1}{2}$, therefore $t_3=8$. Thus the efficiency of this GE is equal to $t = \frac{3+7+8}{3} = 6$. Notice that a complex GE may contain some more specific GE's (e.g. see figure 3 in page 12 of Aamodt & Plaza, 2004). In this case we only need to calculate the efficiency of the complex GE by considering all its cases, regardless if they belong or not to one or more of the specific GE's contained in it. Finally the efficiency of the system is the mean value of the efficiencies of its GE's.

Note: An alternative approach for the representation of cases in a CBR system is the *category and exemplar model* applied first to the PROTOS system ( Porter & Bareiss, 1986). In this model the case memory is embedded in a network of categories, cases and index pointers. Each case is associated with a category. Finding a case in the case library that matches an input description is done by combining the features of the new problem case into a pointer to the category that shares most of these features. A new case is stored in a category by searching for a matching case and by establishing the appropriate feature indices. The process of calculating the efficiency of a system of such type is analogous to the process described in the above example, the only difference being that one has to work with categories instead of GE's. In a similar way one may calculate the efficiency of systems corresponding to other case memory models including Rissland (1983) and Ashley's HYPO system in which cases are grouped under a set of domain-specific dimensions, the MBR model of Stanfill & Waltz (1988), designed for parallel computation rather than knowledge-based matching, etc.

**References**

Aamodt, A. & Plaza, E. (1994), Case-Based Reasoning:: Foundational Issues, Methodological Variations, and System Approaches, *A. I. Communications*, 7, no. 1, 39-52.

Kemeny, J. & Snell, J. l. (1976), *Finite Markov Chains*, Springer-Verlag, New York.
Kolodner, J. (1983), Reconstructive Memory: A Computer Model, *Cognitive Science*, 7, 281-328.

Lebowitz, M. (1983), Memory-Based Parsing, *Artificial Intelligence*, 21, 363-404.

Lei, Y., Peng, Y., Ruan, X. (2001), Applying Case-Based Reasoning to cold forcing process planning, *Journal of Materials Processing Technology,* 112, 12-16 (2001)

Porter, B. & Bareiss, B. (1986), PROTOS: An experiment in knowledge acquisition for heuristic classification tasks, In *Proceedings of the 1$^{st}$ International Meeting on Advances in Learning (IMAL),* 159-174, Les Arcs, France.

Slade, S. (1991), Case-Based Reasoning: A Research Paradigm, *Artificial Intelligence Magazine*, 12/1, 42-55 .

Rissland, E. (1983), Examples in legal reasoning: Legal hypotheticals, In *Proceedings of the Eight International Joint Conference on Artificial Intelligence (IJCAI),* Karlsruhe

Schank, R. (1982), *Dynamic memory; A theory of reminding and learning in computers and people*, Cambridge Univ. Press.

Stanfill, C. & Waltz, D. (1988), The memory-based reasoning paradigm, In *Case-based reasoning, Proceedings from a workshop*, 414-424, Morgan Kaufmann Publ., Clearwater Beach, Florida.

Voskoglou, M. Gr. (2007), A stochastic model for the modelling process, In *C. Chaines et al (Eds), Mathematical Modelling: Education, Engineering and Economics (ICTMA 12)*, 149-157, Horwood Publ. , Chichester.

Voskoglou, M. Gr. (2008), Case-Based Reasoning: A recent theory for problem-solving and learning for computers and people, *Communications in Computer and Information Science (WSKS 08)*, 19, 314-319, Springer-Verlag, Berlin-Heidelberg.